\documentclass[letterpaper, 10 pt, conference]{ieeeconf}  
\IEEEoverridecommandlockouts 
\usepackage{textcomp}
\usepackage{cite}
\usepackage{float}
\usepackage{amsmath,amssymb,amsfonts}
\usepackage{graphicx}
\usepackage{textcomp}
\usepackage{algorithmic}
\usepackage{xcolor}
\def\BibTeX{{\rm B\kern-.05em{\sc i\kern-.025em b}\kern-.08em
    T\kern-.1667em\lower.7ex\hbox{E}\kern-.125emX}}
\usepackage{cite}
\usepackage{comment}
\usepackage{siunitx}
\usepackage{soul}
\usepackage{subcaption}
\usepackage{siunitx}
\usepackage[font=footnotesize]{caption}
\usepackage{hyperref}
\hypersetup{hidelinks}
\usepackage{mathtools}
\usepackage[ruled,vlined,linesnumbered]{algorithm2e} 
\usepackage{wrapfig}

\newcommand{\calS}{\mathcal{S}}

\SetCommentSty{mycommfont}
\usepackage{bm}
\usepackage{mdwmath}
\usepackage{amsmath}

\begin{document}

\title{\LARGE 
Infer and Adapt: Bipedal Locomotion Reward Learning from Demonstrations via Inverse Reinforcement Learning
}
\author{Feiyang Wu, Zhaoyuan Gu, Hanran Wu, Anqi Wu$^\dagger$, and Ye Zhao$^\dagger$
\thanks{F. Wu and A. Wu are with the School of Computational Science and Engineering, Georgia Institute of Technology, Atlanta, GA 30308, USA.
		{\tt\footnotesize \{feiyangwu, anqiwu\}@gatech.edu}}
\thanks{Z. Gu, and Y. Zhao are with the Woodruff School of Mechanical Engineering, Georgia Institute of Technology, Atlanta, GA 30308, USA.
		{\tt\footnotesize \{zgu78, yezhao\}@gatech.edu}}
\thanks{H. Wu is with the College of Computing, Georgia Institute of Technology, Atlanta, GA 30308, USA. {\tt\footnotesize hanran.wu@gatech.edu}}
\thanks{$\dagger$ Co-senior authorship}
}

\maketitle

\begin{abstract} 
Enabling bipedal walking robots to learn how to maneuver over highly uneven, dynamically changing terrains is challenging due to the complexity of robot dynamics and interacted environments. Recent advancements in learning from demonstrations have shown promising results for robot learning in complex environments. While imitation learning of expert policies has been well-explored, the study of learning expert reward functions is largely under-explored in legged locomotion. This paper brings state-of-the-art Inverse Reinforcement Learning (IRL) techniques to solving bipedal locomotion problems over complex terrains. We propose algorithms for learning expert reward functions, and we subsequently analyze the learned functions. Through nonlinear function approximation, we uncover meaningful insights into the expert's locomotion strategies. Furthermore, we empirically demonstrate that training a bipedal locomotion policy with the inferred reward functions enhances its walking performance on unseen terrains, highlighting the adaptability offered by reward learning.

\end{abstract}

\section{Introduction}
\begin{figure*}
    \centering
    \vspace{0.1in}
    \includegraphics[width=0.9\textwidth]{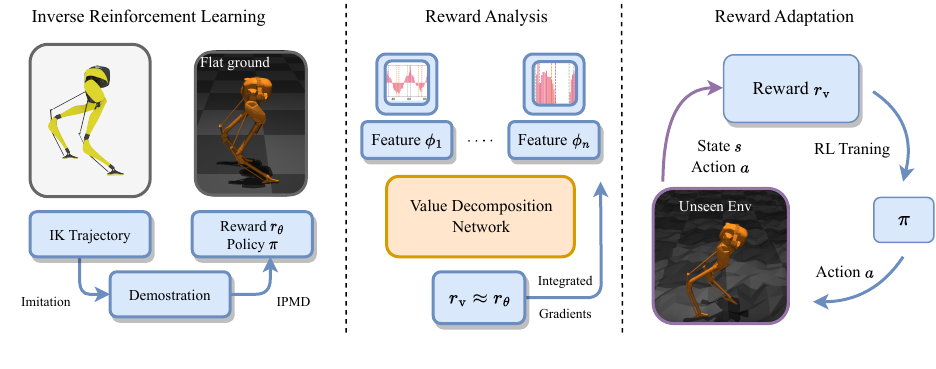}
    \vspace{-0.2in}
    \caption{In this work, we investigate the reward function learned by Inverse Reinforcement Learning algorithms. We propose a two-stage training algorithm for Cassie to learn reward functions and optimal policies from demonstrations. We then analyze the reward function learned from those demonstrations. The learned reward is further used to train RL agents in difficult environments. }
    \label{fig:enter-label}
\vspace{-0.2in}
\end{figure*}

Humans exhibit a remarkable ability to achieve and generalize locomotion strategies from expert demonstrations. This inference ability enables the knowledge transfer from simple tasks to novel tasks and the efficient acquisition of new locomotion skills \cite{van2002obstacle, reynolds2004moving, farmer2018functions, mariscal2022context}. Despite this amazing ability inherent in the human brain, our understanding remains limited regarding the internal representation of a locomotion skill and more importantly, the mechanism for applying acquired skills to novel tasks. Inspired by human's ability to learn from expert demonstrations, this study takes an initial step to mimic this learning ability in the context of bipedal robot locomotion. Moreover, we seek the explainability of the learned skills and demonstrate their generalizability by subjecting the robot to maneuver over various rough terrains.



Imitation learning has been extensively explored as a methodology for learning from demonstration \cite{schaal1999imitation, Ravichandar2020, peng2021amp, florence2022implicit}. Although unable to infer the true intention behind the demonstrations, imitation learning often adopts Reinforcement Learning (RL) formulations to sidestep the problem of lacking an accurate reward function. This RL-based approach requires only designing a reward for tracking the demonstrated actions. 
The development of efficient RL algorithms facilitated a wide range of successful applications of imitation learning for agile bipedal locomotion, such as 
running \cite{siekmann2021sim}, jumping \cite{lirobust2023}, climbing stairs \cite{Siekmann2021Stair}, playing soccer \cite{haarnoja2023learning}, carrying loads \cite{Dao2022Load}, and walking over diverse terrains \cite{krishna2022linear}. 
However, a majority of these works still adopt handcrafted reward functions that heavily rely on domain knowledge and experience. Such reward functions are often tailored for specific environments and have a combination of specific features from the robot's state. Consequently, agents learned under such rewards often lack generalizability and struggle to adapt to new environments.
Inverse Reinforcement Learning (IRL) \cite{ng2000algorithms, arora2021survey}, on the other hand, subsumes the aforementioned imitation learning problem. IRL not only recovers the expert's policy but also the underlying reward function, which captures the essence of the expert's intention and enables adaptations of the robot's motion to unseen tasks. 
Therefore, IRL has gained considerable interest within the robotics community \cite{wulfmeier2017large, liu2022inferring, gan2022energy, chen2023fast }, with some studies employing IRL to gain a deep understanding of the reward function.

However, prior IRL works often presuppose a predetermined feature space and reward structure \cite{wulfmeier2017addressing, gan2022energy}. 
This constrains the expressiveness of reward modeling and leads to limited performance in estimating the true reward functions. Furthermore, the existing robotics IRL works do not analyze the learned reward functions for further usage in practice such as adapting the learned reward for RL during challenging unseen tasks. It remains unclear how one can leverage and transfer the information learned from the reward functions in new environments. 
Moreover, computational complexity has been a hurdle for IRL methods to be widely adopted in the robotics learning community. Recent advances focus on accelerating algorithm efficiency of IRL \cite{garg2021iq,ni2021f,zeng2022maximum, wu2023inverse}. 

In this paper, we develop a novel framework of reward learning, interpretation, and adaptation (Fig.~\ref{fig:enter-label}) to address the aforementioned issues of the existing robotics IRL works. During the learning phase, we employ the Inverse Policy Mirror Descent (IPMD) method \cite{wu2023inverse} to infer the reward from demonstrations. IPMD has been shown to be computationally efficient. It solves the IRL problem with a novel average-reward criterion under a Maximum Entropy framework \cite{ziebart2008maximum, ziebart2010modeling}. The Maximum Entropy framework can discern the most accurate reward estimation by guiding the policy search with the maximum entropy principle. The average-reward criterion also helps to accurately identify reward by dropping the discounted factor that is often used under the classic discounted-reward setting. Since demonstrations often lack an explicit discount factor, using a mismatching discounted factor from the ground truth will lead to drastically erroneous reward function estimations under the discounted setting \cite{wu2023inverse}. Moreover, the average-reward criterion has been thoroughly investigated in the literature and has also been adopted in robotics learning tasks \cite{puterman2014markov, zhang2021average, jin2021towards, li2022stochastic, peters2003reinforcement, ouyang2021adaptive, zhang2021average}.
It has become a common practice for RL benchmarks to use an average-reward metric for evaluation, which further motivates the adoption of the average-reward criterion for solving locomotion tasks.

To gain an in-depth understanding of the learned reward, we employ a  Value Decomposition Network (VDN) \cite{sunehag2017value} and utilize Integrated Gradients (IG) \cite{sundararajan2017axiomatic} to obtain meaningful knowledge of locomotion features leading to high rewards. We will then incorporate such important features into reward design for locomotion in challenging unseen environments, which we refer to as reward adaptation. Note that it is not a new topic to adapt motor and locomotion skills learned from human demonstrations to robots \cite{pastor2009learning, peng2022ase, yang2023learning} or from simulated environments to real-life environments \cite{wulfmeier2017addressing, zhao2020sim, peng2021amp}. However, these works require a sophisticated design and learning of policies or controllers to achieve robust adaptation. Instead, we investigate the possibility of adapting reward functions.
Related methods in adapting reward \cite{nassour2012qualitative, huang2022reward, kaymak2023development}
require crafting intricate, domain-specific reward functions and learning those reward functions under diverse environments to promote the robustness of the policy.
In this work, we use IRL to learn a free-form reward function parametrized by a neural network with inputs directly from the robot's state and action space. 
We show that the learned reward functions contain transferable information about robot locomotion behaviors and verify such properties by training agents using the learned rewards in diverse challenging environments that are not previously seen.
We observe a significant performance boost in walking speed and robustness by incorporating such information. To the best of our knowledge, we are the first to analyze and adapt free-form rewards in a principled way.

The salient contributions of our work are listed as follows:
\begin{itemize}
\item \textbf{Inverse Reinforcement Learning for Bipedal Locomotion}:
We propose a two-stage IRL paradigm to address bipedal locomotion tasks via IPMD. 
In stage one, we obtain expert policies from a fully-body inverse kinematics function of Cassie. In the next stage, IPMD learns reward functions from the near-optimal demonstrations generated by the policies learned in the first stage. 
Our work is the first study that applies IRL to bipedal locomotion under the average-reward criterion.
\item \textbf{Importance Analysis of Expert Reward Function}:
We employ a Value Decomposition Network (VDN) to approximate the inferred locomotion reward function and Integrated Gradients (IG) to analyze the VDN for reward interpretation. By ensuring the monotonicity of the feature space, VDN enables the interpretation of the reward function with IG while preserving model expressiveness. We successfully perform a rigorous analysis of the importance of individual features, 
exposing components of the locomotion behavior that are crucial to its reward functions, thereby guiding the design of new rewards for new environments.
\item \textbf{Reward Adaptation in Challenging Locomotion Environments}:
We further verify that the learned reward from a flat terrain and the important features extracted from our reward analysis can be seamlessly adapted to novel, unseen terrains.
Our empirical results substantiate that the inferred reward function encapsulates knowledge highly relevant to robotic motions that are generalizable across different terrain scenarios.
\end{itemize}
\begin{figure*}[htbp]
    \centering
    \includegraphics[width=\textwidth]{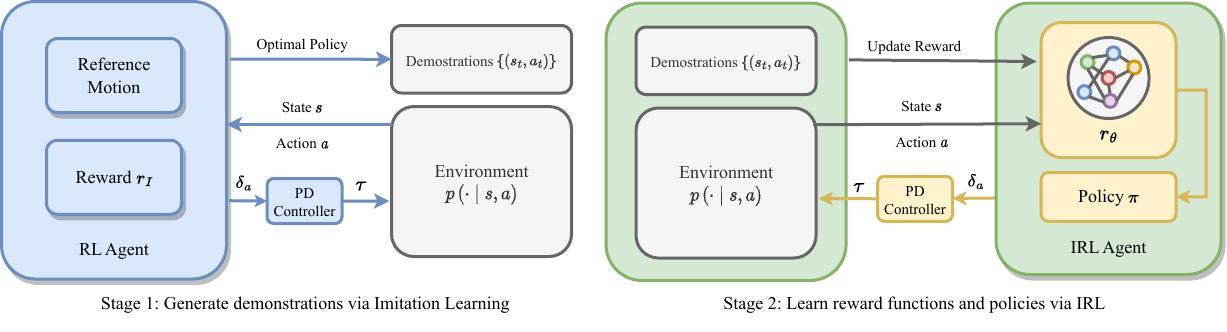}
    \vspace{-0.3in}
    \caption{Our two-stage training pipeline. The blue box denotes the imitation learning part (first stage). The agent is then used to generate expert demonstrations, which are used by the second stage to update the reward and policy using Inverse Policy Mirror Descent.}
    \label{fig:two_stage_training}
    \vspace{-0.1in}
\end{figure*}
\section{Background}
In this section, we introduce preliminaries for Average-reward Markov Decision Processes (AMDPs).
An AMDP is formalized by a tuple $(\mathcal{S},\mathcal{A},\mathsf{P},r)$, where \( \mathcal{S} \) signifies the state space, \( \mathcal{A} \) represents the action space, \( \mathsf{P} \) denotes the transition probability, and \( r \) is the reward function. At each time instance \( t \), the agent selects an action \( a \in \mathcal{A} \) from the current state \( s \in \mathcal{S} \). The system then transitions to a subsequent state \( s' \in \mathcal{S} \) based on the probability \( \mathsf{P}(s'|s, a) \), while the agent accrues an instantaneous reward \( r(s, a) \).

The primary objective of the agent is to establish a policy \( \pi: \mathcal{S} \rightarrow \mathcal{A} \) that optimizes the long-term average reward, mathematically given by

\begin{equation}\label{eq:average-reward}
    \rho^{\pi}(s) := \lim_{{T \to \infty}} \frac{1}{T} \mathbb{E}\left[ \sum_{{t=0}}^{T-1} r(s_t, a_t) | s_0 = s \right].
\end{equation}
Given an expert demonstration set \( \{(s_i, a_i)\}_{i \geq 1} \), IRL aims to extract a reward function that most accurately captures the behavior of the expert. Particularly, in this work, we adopt the Maximum Entropy Inverse Reinforcement Learning (MaxEnt-IRL) framework \cite{ziebart2008maximum}. 

We denote $r_\theta$ as the estimation of the reward function, where $\theta$ is the parameter of the model of choice to represent the reward function $r(s_t, a_t)$ in Eq.~(\ref{eq:average-reward}). For example, $\theta$ can be the weights and biases in a neural network that parameterize the reward. 

In this work, we adopt the environment designed in \cite{xie2018feedback} with the robot's joint-space state as the state space: for any state $s=(x,\hat{x}) \in \calS$, let ${x} = ({q}, {\dot{q}} ) \in \mathbb{R}^{2N}$ represent the robot joint position and velocity, $N = 14$ be the number of joints of Cassie and $\hat{{x}}\in \mathbb{R}^{2N}$ represent the reference motions. Given a reference action $\hat{{a}}$ at a reference state $\hat{{x}}$, the policy outputs an augmentation term $\delta {{a}}$ that corrects the reference action, where $\hat{{a}}, \delta {{a}} \in \mathbb{R}^{M}, M = 10$. The result is a Proportional Derivative (PD) target, ${{a}}=\delta {{a}} + \hat{{a}}$, for a low-level PD controller, which generates a torque ${\tau} \in \mathbb{R}^{M}$ to track joint angles. 

\section{Methods}
In this section, we first introduce the pipeline that applies Inverse Policy Mirror Descent (IPMD) for bipedal locomotion to learn reward functions. We then outline our approach to analyze the learned reward function and methodology of conducting reward adaptation experiments.

\subsection{Two-Stage Learning Pipeline}

Recent RL techniques for bipedal locomotion rely on carefully constructing the state and action space and designing sophisticated reward functions \cite{xie2018feedback, li2021reinforcement, chen2020optimal}. 
IRL models endow capabilities to learn from demonstrations. However, a practical challenge often arises: \textit{what type of trajectory data should IRL leverage for effective learning?} Directly recording trajectories from robots such as motion capture approaches can be laborious and time-consuming, while data derived from model-based methods such as inverse kinematics or trajectory optimization often suffer from inaccurate models and unrealistic assumptions. 
To get high-quality demonstrations for effective IRL, we will use imitation learning with the Markov Decision Process (MDP) environment similar to \cite{xie2018feedback}, which can produce computationally convenient and dynamically accurate expert demonstrations, even if we only have trajectory data generated by model-based methods.

Accordingly, we propose a two-stage IRL learning pipeline that utilizes both imitation learning and IPMD.
Our approach is graphically summarized in Fig.~\ref{fig:two_stage_training}. 
In the first stage, we apply imitation learning on data generated via inverse kinematics to create near-optimal demonstrations, as subsequent IRL training and reward analysis require dynamically accurate demonstrations. The imitation learning style reward function $r_I$ used in this environment is defined as a weighted sum of tracking rewards at the joint level:
\begin{align}
    r_{I} &= c_1 e^{-E_{\text{joint}}} + c_2 e^{-\
    |p_{\rm CoM}-p_{\rm CoM}^r\|} + c_3 e^{-\|p_o-p_o^r\|}
\end{align}
where $c_1, c_2, c_3$ are constant coefficients, $E_{\text{joint}}$ is a weighted Euclidean norm of the difference between the current joint position ${q}$ and the reference joint position ${q}^r$: $E_{\text{joint}}^2:= {w}^T ({q}-{q}^r)^2$, $ {w}, {q}, {q}^r \in \mathbb{R}^N$. $p_{\rm CoM}$ denotes the Center of Mass (CoM) position, and $p_o$ denotes pelvis orientation. The superscript $r$ denotes the reference motion.

Using expert demonstrations generated from the first stage, the second stage employs our IPMD method to learn both the optimal policy and the associated reward function in the form of a deep neural network.
Concretely, in each iteration of the IPMD algorithm, we sample state-action pairs by interacting with the environment and also sample state-action pairs from demonstrations. We then employ Temporal-Difference (TD) to evaluate our current policy given the first set of sampled pairs from the environment and apply a Mirror Descent step to improve the current policy. At the end of the iteration, we update the reward estimation through gradient descent given the two sets of sampled pairs. Due to the space limit, more details can be referred to in \cite{wu2023inverse}.
\subsection{Analysis of the Learned Reward Function}
We extend our study to a detailed analysis of the learned reward function. The reward function $r_\theta$ is a deep neural network that inherently lacks interpretability due to its black-box nature.
To tackle this issue, we employ a more interpretable model, Value Decomposition Network (VDN) \cite{sunehag2017value}, which approximates the reward function and explains the significance of locomotion features in determining the reward value. 
VDN maintains a monotonic relationship between its input and output by constraining the weights and biases of the network to be positive, ensuring continuous positive gradients \cite{rashid2020monotonic}. This property of VDN allows us to establish a monotonic mapping from the state space to the reward output without compromising the learned reward's accuracy due to its usage of neural networks \cite{rashid2020monotonic}.

\begin{figure}[t]
  \begin{center}
    \includegraphics[width=0.4\textwidth]{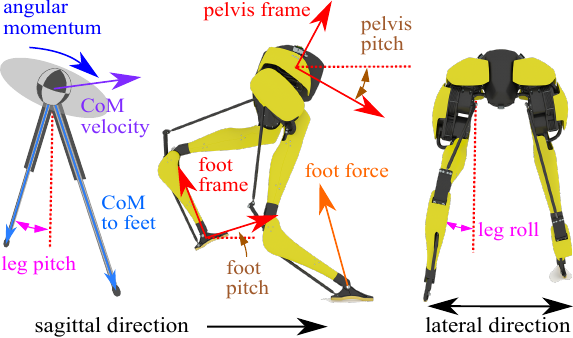}
  \end{center}
\vspace{-0.1in}
\caption{Illustration of important features for Cassie locomotion.}
\label{fig:feature}
\end{figure}

Additionally, we aim to explore the features that are highly relevant to bipedal locomotion but may not be directly present in the state space, such as the leg length or ground reaction force, to study how these indirectly observed features affect the reward function. To facilitate this, we extend the input space of our approximation model to include these features. The full list of selected features is in Table \ref{table:features} and a majority of them are annotated in Fig.~\ref{fig:feature}. Through this approximation, we establish a relationship between the selected features and the reward function, while keeping the IRL training process separate and intact, allowing it to preserve the expressive power of deep neural nets.
\begin{table}
\renewcommand{\arraystretch}{1.3}
\caption{Considered Features for Approximating Learned Rewards}
\label{table:features}
\centering
\begin{tabular}{|c|c|c|}
        \hline
        state & action & Euclidean norm of action\\
        \hline
        leg roll  & leg pitch & pelvis pitch\\
        \hline
        hip yaw & foot pitch & foot force\\
        \hline
        CoM velocity& CoM angular momentum & CoM to center of foot \\
        \hline
    \end{tabular}
\vspace{-0.2in}
\end{table}

Equipped with an interpretable approximation from VDN, we proceed to further dissect the learned reward function using a set of neural network interpretation techniques. In particular, we find Integrated Gradients (IG), a widely recognized tool in the Deep Learning community, to be highly suitable for our objectives \cite{sundararajan2017axiomatic}. IG allows us to analyze the effect of individual features on the overall landscape of the reward function by perturbing the input and observing the resulting gradient changes, which in our case are manifested as variations in the neural network weights. 
We also find that directly applying IG to the original reward function itself does not yield any meaningful outcome, due to the highly nonlinear relationship between the input (states and actions) and the output (rewards). This validates the necessity of using VDN to approximate the original reward function for better reward interpretation with IG. 

\subsection{Adaptability of the Learned Rewards on Difficult Terrains}

In this context, we explore whether our learned reward function harbors generalized knowledge that enables adaptability across varying terrains. Specifically, we test its efficacy in a purely RL-driven training paradigm, without the need for additional expert demonstrations. 
Intriguingly, the RL guided by the learned reward not only allows training from scratch but also produces a better performance compared to policies learned from the hand-crafted reward. Even though the reward function was originally trained on flat terrain, our learned reward successfully guides the agent's learning in more complex environments.

This observation aligns well with the intuition that a well-designed reward function encapsulates generalizable environmental knowledge. To validate this point, we present results showcasing Cassie's capability to navigate difficult terrains. 

More interestingly, with the understanding of reward functions, we show that factored components inside the reward function, i.e., those found during our reward function analysis, can improve the quality of locomotion behaviors. 
This constitutes a significant contribution to the field, as traditional algorithms often require the crafting of intricate, domain-specific reward functions.

\section{Experiments}
\subsection{Two-Stage Learning Setup}
Our experiments of Cassie locomotion were conducted using the MuJoCo physics simulator~\cite{Mujoco}. The training pipeline consists of two main stages as illustrated in Fig.~\ref{fig:two_stage_training}.

\subsubsection{First Stage -- Training the Imitation Agent}
We train the Imitation agent using Soft Actor-Critic (SAC)~\cite{haarnoja2018soft}. 
The discount factor $\gamma$ for this stage is set to $0.99$. Both the policy and value functions are parameterized by $256 \times 256$ Multi-Layer Perceptrons (MLPs). For implementation, we adopt the state-of-the-art codebase from stable-baselines3~\cite{stable-baselines3}.

\subsubsection{Second Stage -- Learning reward functions and policies via IRL}
We use the Inverse Policy Mirror Descent (IPMD) method described in~\cite{wu2023inverse}. The reward function, policy, and value functions are all represented by $256 \times 256$ MLPs. 

\subsubsection{Training Parameters}
Both agents are trained using $5 \times 10^6$ samples. We employ an experience replay buffer with a capacity of $1 \times 10^6$ and utilize a batch size of $512$. The Adam optimizer~\cite{KingBa15} is employed with a learning rate set at $3 \times 10^{-4}$.
These parameter settings are consistent with established norms for training Deep RL algorithms.

From a simulation experiment, the optimal expert agent obtained an episodic reward of $447.2$ while generating the corresponding expert demonstration data for the second stage; the IRL agent trained with IPMD reached a better performance---an episodic reward of $482.87$. The fact that the IRL agent outperforms the expert demonstrations reflects the superiority of our methodology. The qualitative performance of the IRL agent has no distinguishable difference compared to the imitation agent, this is surprising since we learn both the reward functions and policies from scratch, while in the imitation learning case, a complicated reward function has already been established. 

\subsection{Reward Analysis}
\begin{figure}
    \centering
        \centering
        \includegraphics[width=0.48\textwidth]{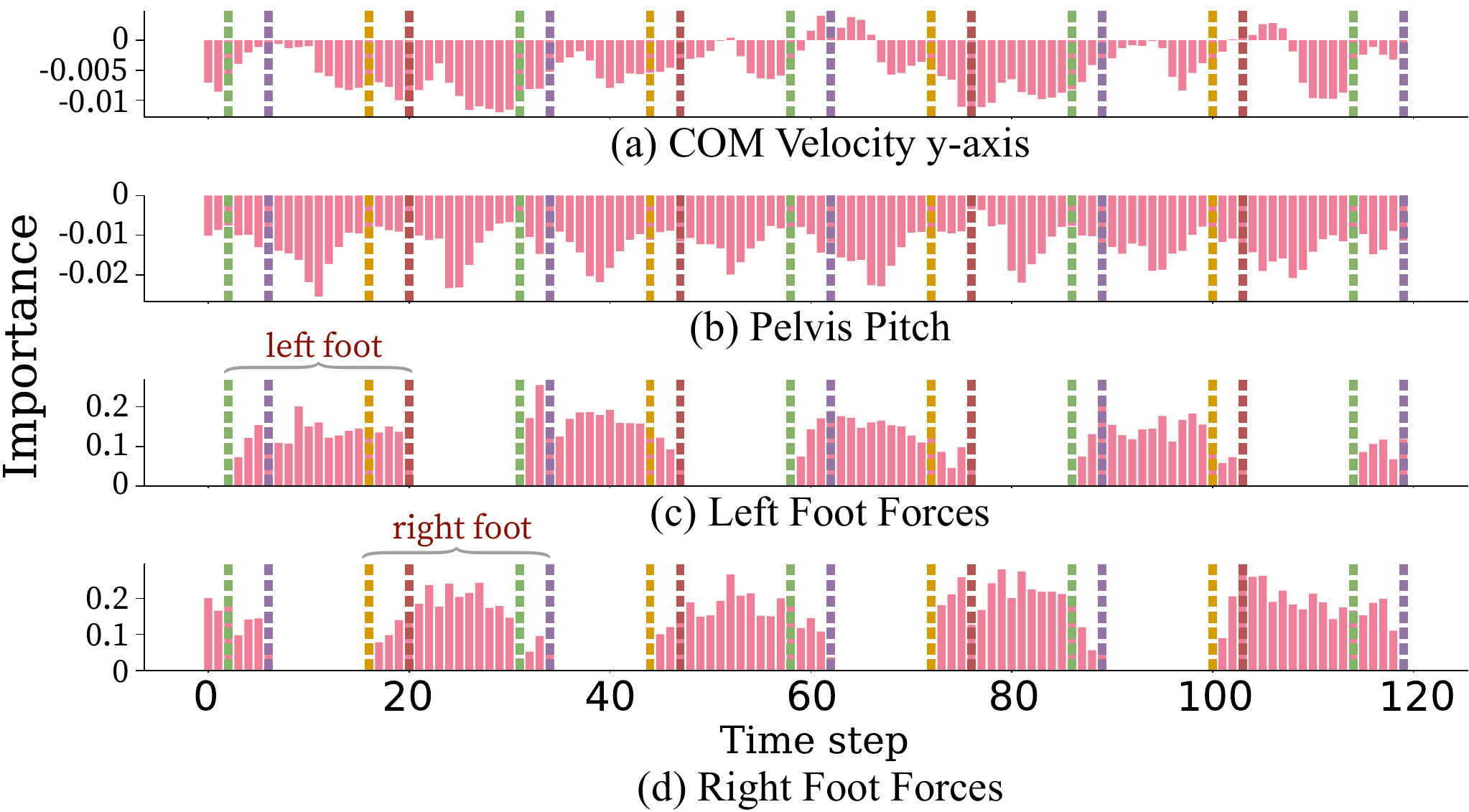}
          \vspace{-0.2in}
      \caption{The top four most important features: CoM lateral velocity, pelvis pitch angle, left and right foot forces. The vertical dashed lines represent time steps when the foot touches and leaves the ground. Green indicates when the left foot strikes and red is for the left foot taking off from the ground. The same for orange (strike) and purple (take off) for the right foot.  }
    \label{fig:reward_inference_results}
\vspace{-0.2in}
\end{figure}

For the Value Decomposition Network (VDN), we adhere to the same network structure as described in~\cite{rashid2020monotonic}. We gather training samples by recording the states of the Cassie robot, along with additional data necessary for computing the features of interest. We list all features we find worth investigating in Table \ref{table:features}.
As we aim to approximate the learned reward function, we use the rewards generated by $r_\theta$ as regression targets for the VDN. The optimization objective is the Mean Squared Error (MSE), thereby transforming the training of VDN into the following optimization problem:
$\min_{\psi} \; \text{MSE}(\text{VDN}(\psi), r_\theta),$
where $r_\theta$ is the learned reward function and $\psi$ represents the parameters of the VDN, i.e., the weights and biases in neural networks. We record and compute specified feature data as input, and collect rewards computed from those data using the learned reward functions as regression targets.
We employ the Adam optimizer with a learning rate of $3 \times 10^{-4}$ to train the VDN. 
To interpret the contribution of each feature to the reward function, we employ Integrated Gradients (IG)~\cite{sundararajan2017axiomatic}, which is further implemented by Captum~\cite{kokhlikyan2020captum}. 
Fig.~\ref{fig:reward_inference_results} demonstrates that the reward function approximated by the VDN aligns well with our intuitive understanding of what features are  important for bipedal locomotion. 
We plot the importance change of four features to the reward during one typical Cassie walking motion executed by the IRL agent. 

We find that some features of interest exhibit periodic patterns, due to the nature of the periodic walking motion. This aligns with our understanding of bipedal locomotion. 
Some particular features exhibit a strong influence on the reward even if they have no particular pattern. We note that pelvis pitch, plotted in Fig.~\ref{fig:reward_inference_results}, has significant values compared to its small-scale raw input data.
We conjecture that the pelvis pitch plays an important role in maintaining the stability of the robot during walking. 
Other features also have strong correlations with their physical meaning. For example, the left foot has ground reaction force only when it is in contact with the ground. This is rather intuitive for robot locomotion. 

\subsection{Adaptive Reward Function}
\begin{figure}
    \centering
    \vspace{0.1in}
    \begin{subfigure}[b]{0.2\textwidth}
        \centering
        \includegraphics[width=\textwidth]{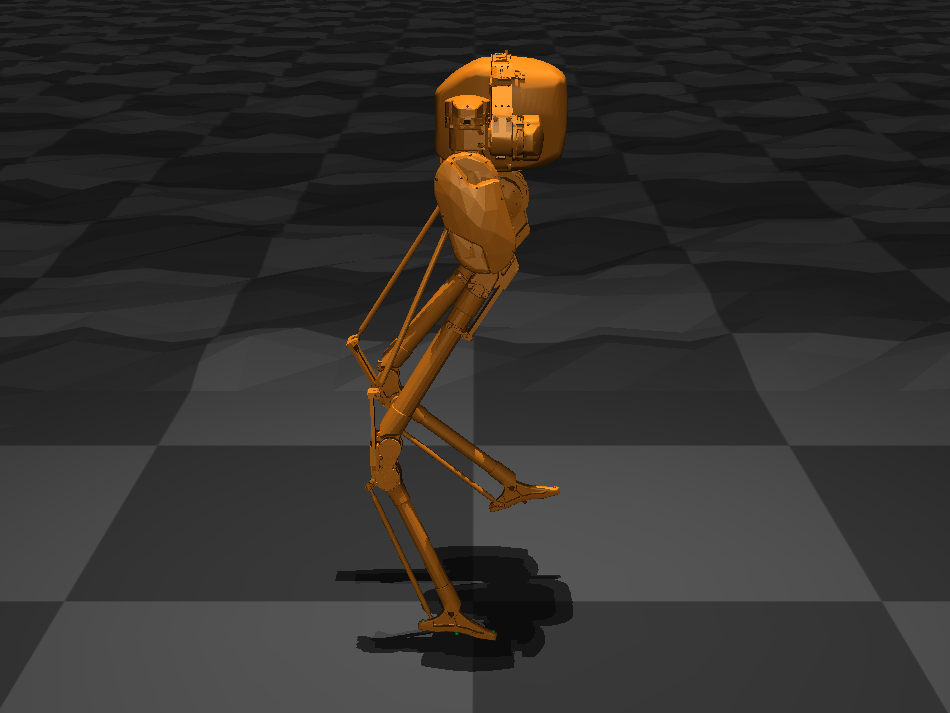}
    \caption{Random Perturbed Terrain}
        \label{fig:terrain_random}
    \end{subfigure}
      \vspace{0.07in}
  \begin{subfigure}[b]{0.2\textwidth}
        \centering
        \includegraphics[width=\textwidth]{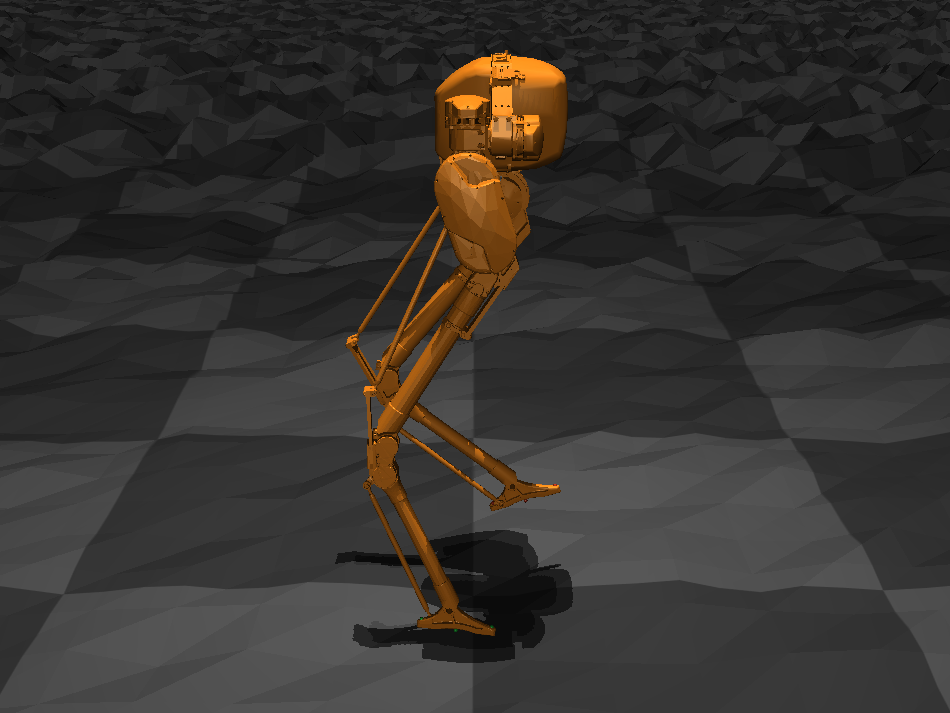}
        \caption{Gradually Perturbed Terrain}
        \label{fig:terrain_random_gradual}
    \end{subfigure}
    \begin{subfigure}[b]{0.2\textwidth}
        \centering
        \includegraphics[width=\textwidth]{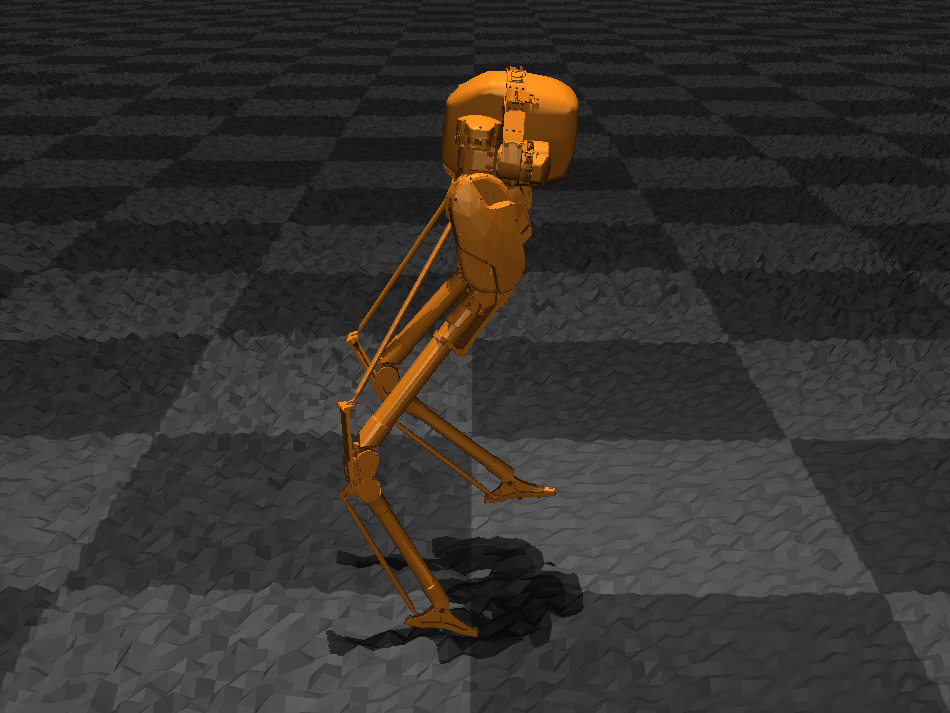}
        \caption{Gravel Terrain}
        \label{fig:terrain_gravel}
    \end{subfigure}
    \begin{subfigure}[b]{0.2\textwidth}
        \centering
        \includegraphics[width=\textwidth]{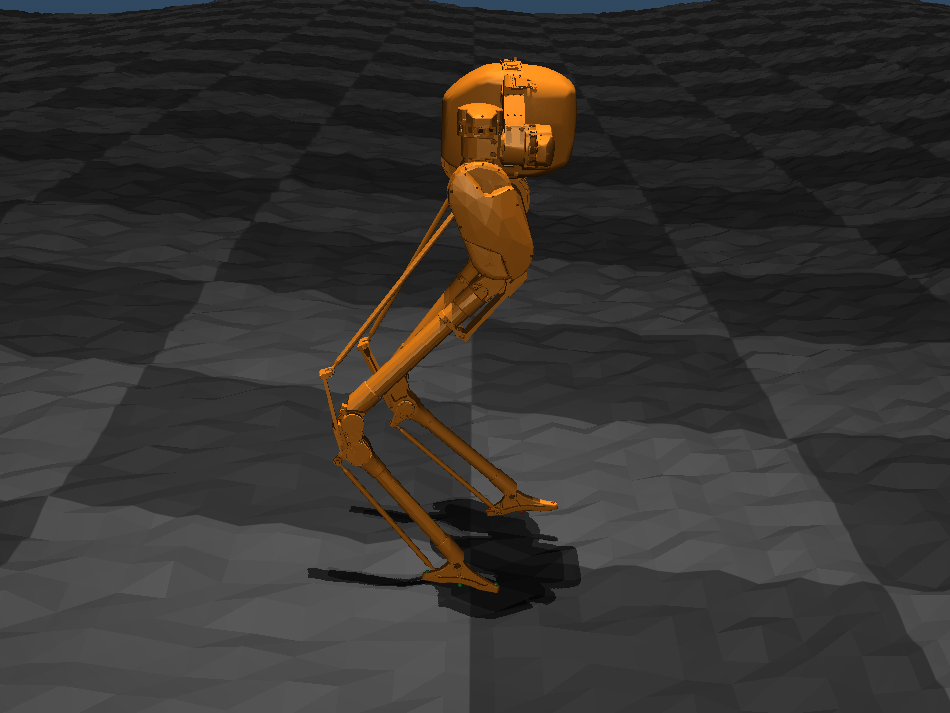}
        \caption{Sine Wave Terrain}
        \label{fig:terrain_sine_wave}
    \end{subfigure}
    \caption{Random terrains generated for testing the learned reward function.}
    \label{fig:terrain_showcase}
\end{figure}

We generate a variety of uneven terrains in MuJoCo environments as shown in Fig.~\ref{fig:terrain_showcase}. In particular, we create (a) random perturbed terrain, (b) gradually perturbed terrain, (c) gravel terrain, and (d) sine wave terrain, each with maximum height capped at 0.2, 0.3, 0.1, 0.4 meters respectively. 
These categories serve to evaluate the adaptability and generalization capacity of our learned reward function.

We train the agent from scratch using SAC with a discount factor of \( \gamma = 0.99 \), following the same setup as in our imitation learning model. For comparative analysis, we also train a baseline RL agent with a handcrafted reward function defined as
$r_h = r_{f} + r_{s} - r_{c},$
where \( r_f \) encourages forward movement and corresponds to the sagittal velocity; \( r_s \) is a locomotion survival reward, awarded when Cassie torso remains upright; and \( r_c \), the control cost, is defined as \( r_c = \|a\|_2 \).

\begin{figure}
        \centering
        \includegraphics[width=0.45\textwidth]{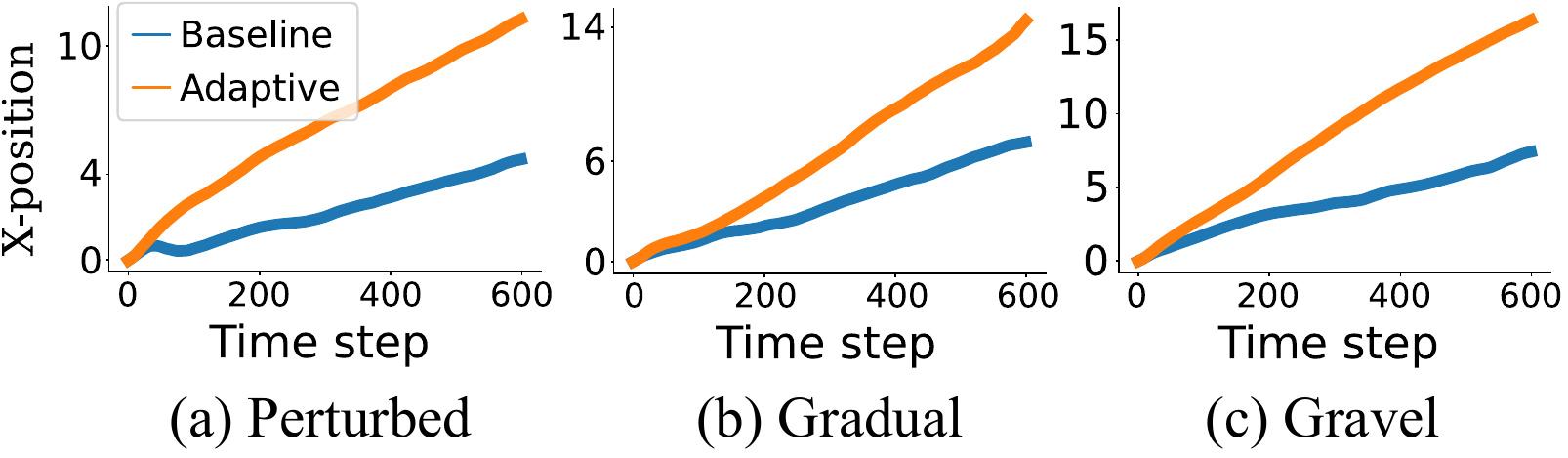}
       \caption{Sagittal travel distance comparison between baseline model using $r_h$, and adaptive reward model using $r$. Note that even though the Baseline model can walk up to maximum time steps, it can not walk as far as the one using the adaptive reward.}
    \label{fig:com}
    \vspace{-0.1in}
\end{figure}

The baseline agent manages to navigate these terrains, albeit in a less graceful manner with jerky motions (see the submitted video). In contrast, our approach uses a modified reward function:
$r = r_h + r_\theta,$
where \( r_\theta \) is the reward function learned from IRL. We refer to $r$ as the Adaptive reward. We record the average sagittal velocity of CoM when comparing the baseline reward model and the adaptive reward model side by side. The results can be found in Table \ref{table:adaptation_walking_speed}. 
We also plot the sagittal travel distance in each environment, which is shown in Fig.~\ref{fig:com}.
We find that incorporating \( r_\theta \) significantly accelerates learning and produces more natural and robust locomotion behaviors, substantiating the transferability of the learned reward function across domains.
\begin{table}[h]
\renewcommand{\arraystretch}{1.3}
\caption{Average Center of Mass velocity (m/s) in sagittal direction}
\label{table:adaptation_walking_speed}
\centering
\begin{tabular}{|c|c|c|}
        \hline
        Terrain & Baseline & Adaptive \\
        \hline
        Perturbed  & 0.2617 & 0.6249 \\
        \hline
        Gradual & 0.3970 & 0.8015 \\
        \hline
        Gravel & 0.4132 & 0.9106 \\
        \hline
    \end{tabular}
\vspace{-0.1in}
\end{table}

\subsection{Analysis-based Adaptive Reward Design}
With the adaptive reward, the robot is able to walk on unseen rough terrains. However, instances of undesirable walking gait still occasionally occur. Specifically, using the adaptive reward alone, Cassie's CoM exhibits a higher sagittal CoM velocity. In reality, such behavior is undesirable as this inclination creates instability during locomotion in rough terrains. Consequently, the robot needs to maneuver agilely to maintain balance during walking. This leads to the robot deviating from the original lateral position, which is reflected by large variations of CoM velocity along the lateral direction. 
With the understanding of the learned reward, a natural question arises: \textit{can we further exploit the learned reward functions to shape the locomotion behavior?} We answer this question affirmatively. The top important features uncovered in the Reward Analysis improved the stability of walking behaviors when incorporated with the learned reward.
As such, we incorporate important features discovered from the reward analysis to boost the stability of the robot, or "regularize" the robot's motion.
To do this, we add three additional terms with high importance scores to the adaptive reward: pelvis orientation, pelvis pitch angle, and CoM velocity, which are implemented to follow their reference motions on the flat ground. We denote such rewards as 
$\label{eq:regularization-reward}
    r_v = e^{-\|q_o-q_o^r\|_2} + e^{-\|q_{\text{pitch}}-q_{\text{pitch}}^r\|_2} + e^{-\|v_{\rm CoM}-v_{\rm CoM}^r\|_2},
$
where $q_o$ denotes the pelvis orientation in a quaternion form, $q_{\text{pitch}}$ is the pelvis pitch angle, and $v_{\rm CoM}$ is the CoM velocity.

\begin{figure}
    \centering
        \centering
        \includegraphics[width=0.45\textwidth]{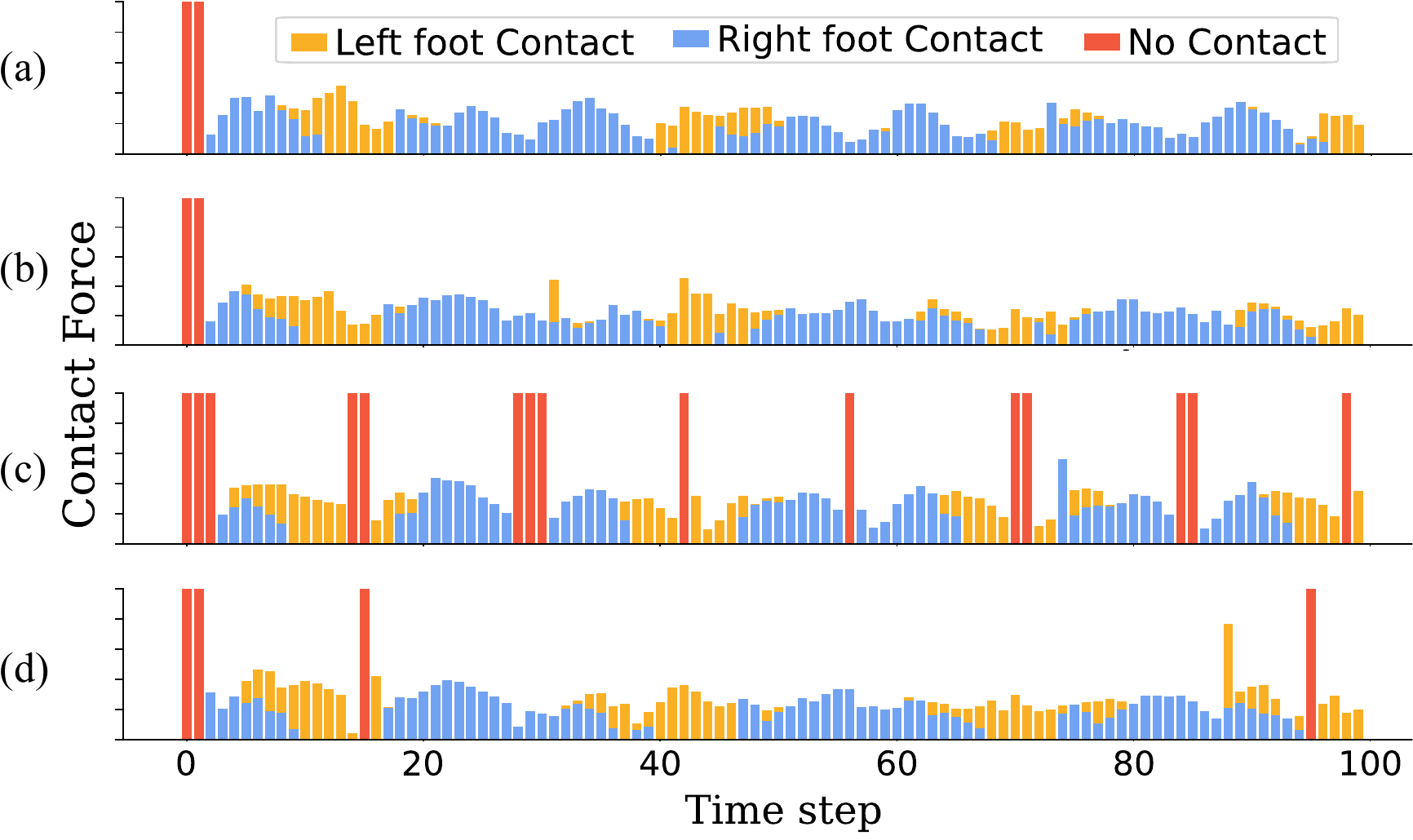}
    \caption{Ground reaction force with four reward setups: (a) $r_h$, (b) $r_h+r_v$, (c) $r_h+r_{\theta}$, (d) $r_h+r_{\theta}+r_v$. The orange bar denotes the left foot force, while the blue the right. The red bar denotes time steps when no ground reaction force exists for either foot.}
    \label{fig:ground_reaction_force}
\vspace{-0.2in}
\end{figure}

To verify the efficacy of the $r_v$, we train RL agents with SAC on four combinations of reward functions: the baseline model $r_h$, the regularized model $r_h+r_v$, the adaptive model $r_h +r_{\theta}$, and the regularized adaptive model $r_h +r_{\theta}+r_v$. 
\begin{figure}[t]
    \centering
        \includegraphics[width=0.48\textwidth]{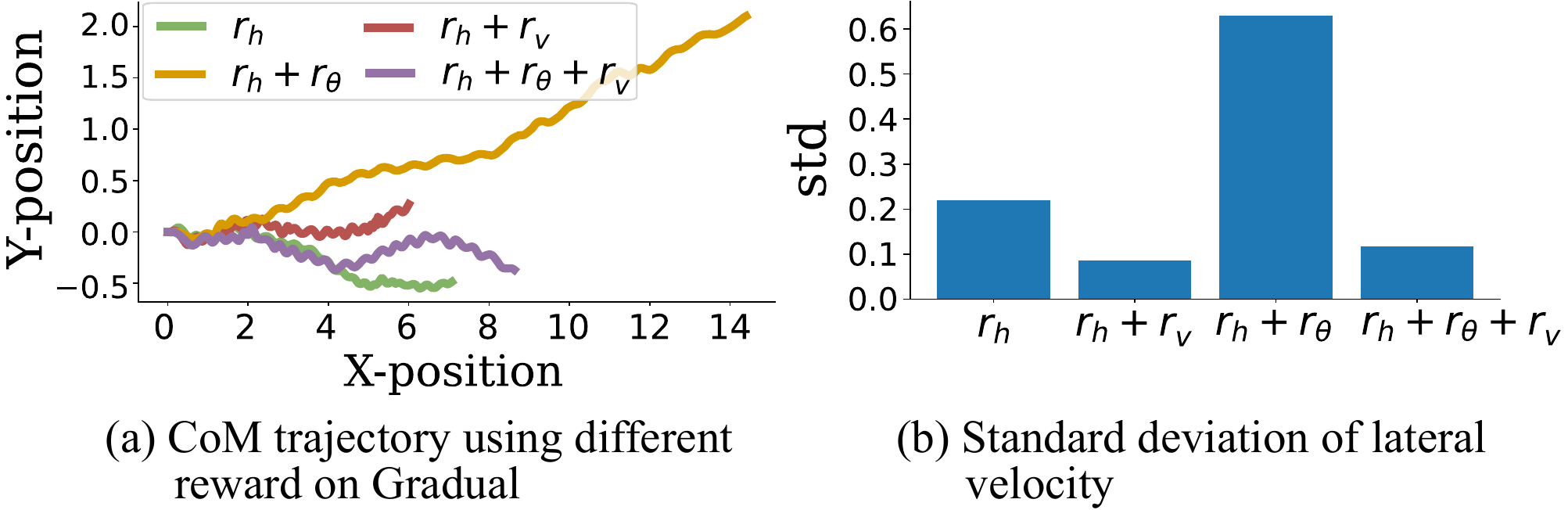}
    \caption{Results for regularizing robotics behavior.}
    \label{fig:regularizer}
\end{figure}
We plot the CoM trajectory, and standard deviation of the velocity drift along the lateral direction in Fig.~\ref{fig:regularizer}. 
Although the adaptive model allows the robot to walk further, it has a higher deviation from its original lateral position and a higher deviation of lateral velocity. We conjecture that this is partially due to the fact that the orientation is less emphasized by the adaptive reward. 
We also observe that purely using the adaptive reward results in a "hopping" behavior where each walking step has a brief flight phase. In reality, such loss of ground contact can lead to a highly unstable walking motion and pose a risk of failure. Surprisingly, the integration of additional regularizing terms in the reward function $r_v$ mitigates such undesirable hopping behaviors. We plot the ground reaction force of all four models in Fig.~\ref{fig:ground_reaction_force}. Time steps when undesirable behaviors (both feet are in the air) occur are annotated with red color bars. 

Fig.~\ref{fig:ground_reaction_force}(b) and (d) show a more stable and natural walking motion, compared with Fig.~\ref{fig:ground_reaction_force}(c) (also shown in the video), indicating the efficacy of the $r_{v}$ reward in regulating the robot's behavior. This result further demonstrates that the augmentation of the reward function with relevant extracted features leads to improved locomotion performance.

\subsection{Zero-Shot Validation}
We observe that agents trained on diverse terrains display enhanced stability when deployed in unseen environments. For example, Cassie is able to navigate sinusoidal terrains with random height variations (Fig.~\ref{fig:terrain_sine_wave}), without additional training. This corroborates the idea that the learned reward embodies a form of generalized knowledge beneficial for robotic locomotion across a range of terrain scenarios.

\section{Conclusion}

In this work, we employ an IRL method to solve bipedal locomotion problems. Our analyses reveal that the learned reward function encapsulates meaningful insights and also serves as a valuable guide to understanding the underlying principles of robotic motion. 
The ability to learn and adapt using the inferred reward function paves the way for new avenues of research in robotics, particularly in the domain of reward inference and environmental adaptability. 
Our work supports the notion that leveraging learned reward functions could substantially accelerate the design, training, and deployment of robotic systems across a myriad of real-world scenarios. Our future direction will focus on hardware implementation on the Cassie robot.

\clearpage 
\bibliographystyle{IEEEtran}
\bibliography{references}

\end{document}